%% file: 00_main.tex
\pdfoutput=1
\documentclass[10pt,twocolumn,letterpaper]{article}
\usepackage[accsupp]{axessibility} 
\usepackage{iccv}
\usepackage{times}
\usepackage{subcaption}
\usepackage{epsfig}
\usepackage{graphicx}
\usepackage{amsmath}
\usepackage{amssymb}
\usepackage{multirow}
\usepackage{booktabs}
\usepackage{algorithm} 
\usepackage{algpseudocode} 
\usepackage{color, colortbl}
\usepackage{listings}
\usepackage{xcolor}
\usepackage{color, colortbl}
\definecolor{rulecolor}{RGB}{0,71,171}
\definecolor{Gray}{gray}{0.9}
\usepackage{listings}
\usepackage{color}

\definecolor{mygreen}{rgb}{0,0.6,0}
\definecolor{mygray}{rgb}{0.5,0.5,0.5}
\definecolor{mymauve}{rgb}{0.58,0,0.82}

\usepackage[
  separate-uncertainty = true,
  multi-part-units = repeat
]{siunitx}

\usepackage[pagebackref=true,breaklinks=true,letterpaper=true,colorlinks,bookmarks=false]{hyperref}

\usepackage{bbm}

\usepackage[capitalize]{cleveref}

\iccvfinalcopy %

\def\iccvPaperID{10202} %

\ificcvfinal\fi

\newcommand{\methodname}{Time-Tuning}
\newcommand{\methodabbrev}{\textsc{TimeT}}

\newcommand{\midsepremove}{\aboverulesep = 0.3mm \belowrulesep = 0.5mm}
\midsepremove

\begin{document}

\title{Time Does Tell: Self-Supervised \textit{Time-Tuning} of Dense Image Representations}

\author{Mohammadreza Salehi, Efstratios Gavves,  Cees G. M. Snoek, Yuki M. Asano\\
QUVA Lab, University of Amsterdam\\
{\tt\small (s.salehidehnavi, e.gavves, c.g.m.snoek, y.m.asano)@uva.nl}
}

\maketitle

\begin{abstract}

Spatially dense self-supervised learning is a rapidly growing problem domain with promising applications for unsupervised segmentation and pretraining for dense downstream tasks.
Despite the abundance of temporal data in the form of videos, this information-rich source has been largely overlooked.
Our paper aims to address this gap by proposing a novel approach that incorporates temporal consistency in dense self-supervised learning.  
While methods designed solely for images face difficulties in achieving even the same performance on videos, our method improves not only the representation quality for videos – but also images. 
Our approach, which we call \textit{time-tuning}, 
starts from image-pretrained models and fine-tunes them with a novel self-supervised temporal-alignment clustering loss on unlabeled videos. 
This effectively facilitates the transfer of high-level information from videos to image representations. 
Time-tuning improves the state-of-the-art by 8-10\% for unsupervised semantic segmentation on videos and matches it for images.
We believe this method paves the way for further self-supervised scaling by leveraging the abundant availability of videos. The implementation can be found here : \url{https://github.com/SMSD75/Timetuning}

\end{abstract}

\input{Figs/Fig1}

\input{01_introduction}
\input{02_related_work}

\input{03_method}

\input{04_experiments}

\input{05_conclusion}

{\small
\bibliographystyle{ieee_fullname}
\bibliography{main}
}
\clearpage
\newpage 
\input{99_appendix}

\end{document}

%% file: Figs/Fig1.tex
\begin{figure}[!t]
    \centering
        \includegraphics[width=1\linewidth,  trim={0.8cm 0cm 1cm 0.7cm},clip]{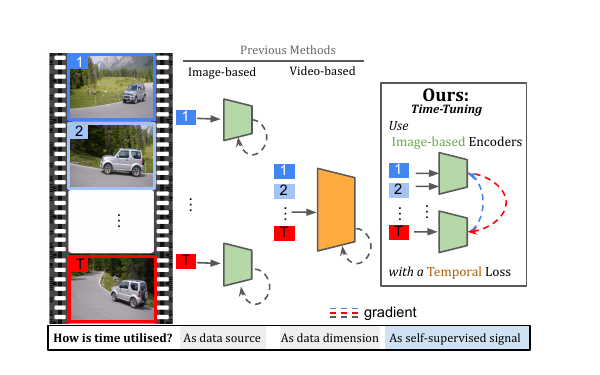}
        \vspace{-3em}
    \caption{\textbf{Time-tuning compared to previous methods.} 
    Unlike existing methods that ignore or utilize expensive 3D models to implicitly model temporal information, the proposed method explicitly incorporates temporal consistency in dense feature representations using a temporal self-supervised loss. The method starts with a 2D encoder pretrained on images and fine-tunes it using unlabeled videos. This approach leads to improved performance not only for videos but also for images.}
\label{fig:fig_1}
\end{figure}

%% file: 01_introduction.tex
\section{Introduction}
Dense self-supervised learning, whereby  meaningful deep features for each pixel or patch of input are learned in an unsupervised manner, has recently received increasing attention~\cite{ziegler2022self, henaff2021efficient, van2021unsupervised, seitzer2022bridging}. 
By learning spatially consistent features for different views of an input, %
strong gains in unsupervised semantic segmentation have been achieved using unlabeled images.
However, so far, an even more information-rich source for unsupervised training has been largely overlooked: videos. 
With their additional time dimension and being the most rapidly growing form of digital content, they are well-suited to scaling dense self-supervised learning even further.

Some efforts have already been made to learn from the video domain by using different frames from a video as augmentations~\cite{fernando2017self, xu2019self, ranasinghe2022self} or by involving temporal correspondence~\cite{zhang2023boosting, wang2022look}; however, they mostly did it in a supervised way ~\cite{hur2016joint, zhu2017deep, gadde2017semantic, liu2017surveillance, xu2018dynamic, li2018low, jain2019accel, hu2020temporally, liu2020efficient, li2022video, sun2022coarse, nirkin2021hyperseg}, which is not scalable specifically for dense tasks where the number of targets can increase significantly as the number of pixels grows. To this end, self-supervised learning approaches offer a solution by reducing the need for supervision. However, these methods typically rely on the notion of ``views'', which involves learning similar features for corresponding locations over time. This usually leads to a chicken-and-egg problem, where the correspondences are required for learning dense features -- which in turn enable good correspondences~\cite{jabri2020space}.

In images, the challenge is trivially solved by considering the correspondence function based on the augmentation function. For instance, In the case of color augmentations, this correspondence is simply given by the identity. However, shifts in time cannot be viewed as mere augmentations. As we demonstrate through a new evaluation protocol, using image models on frames alone is not nearly as effective. A similar finding has also been reported, albeit for non-dense works ~\cite{gordon2020watching, kipf2021conditional}, which assumed the passage of time  as an image augmentation. These works have generally reported reduced performances, even when compared to simple image-based pretraining methods. Similarly, video-level tasks~\cite{feichtenhofer2021large, fernando2017self, thoker2022severe} assume sufficiently similar semantics between different frames. This is also not true for dense tasks, as static features can only be assumed where nothing is moving -- which is rare due to possible object, camera, and background motion between the frames.

To address this challenge, we propose to model the additional time-dimension explicitly to identify which pixels should retain similar embeddings and which should not. We propose two separate modules to tackle the correspondence and the dense learning, respectively.  For the former, we introduce the Feature-Forwarder (\textsc{FF}) module, which breaks the mentioned chicken-and-egg loop by leveraging the good tracking performance of pretrained image models, and allows an approximate second ``view'' that can then be treated as a target for the further dense self-supervised loss. On top of this, we introduce a spatio-temporally dense clustering module, which learns unsupervised clusters across samples, locations and time. Using these two components and starting from image-pretrained features, our proposed method allows \textit{time-tuning} (\methodabbrev) the dense representation in a self-supervised manner, see Figure~\ref{fig:fig_1}.

Finally, we demonstrate that \methodabbrev~ paves the way for further scaling of self-supervised learning by leveraging the abundant availability of video datasets and transferring their knowledge to the image domain. This results in consistently achieving state-of-the-art performances not only for the task of unsupervised semantic segmentation of videos, but also for unsupervised \textit{image} semantic segmentation, a feat previously out of reach for methods trained on videos.

Overall, this paper makes the following contributions:
\begin{itemize}
    \item We show that image-based unsupervised dense segmentation models applied to videos exhibit degraded performance and lack temporal consistency in their segmentation maps.
    \item Building on this observation, we propose a novel dense self-supervised learning method that utilizes temporal consistency as a learning signal.
    \item We demonstrate that our method enables the scaling of self-supervised learning by leveraging abundant video datasets and effectively transferring knowledge to the image domain. Our approach consistently achieves state-of-the-art performance for both images and videos, opening up new opportunities in the field.
    
\end{itemize}

\begin{figure*}[!t]
    \centering
    \includegraphics[width=2.0\columnwidth, trim={0.8cm 0cm 0.2cm 0.2cm},clip]{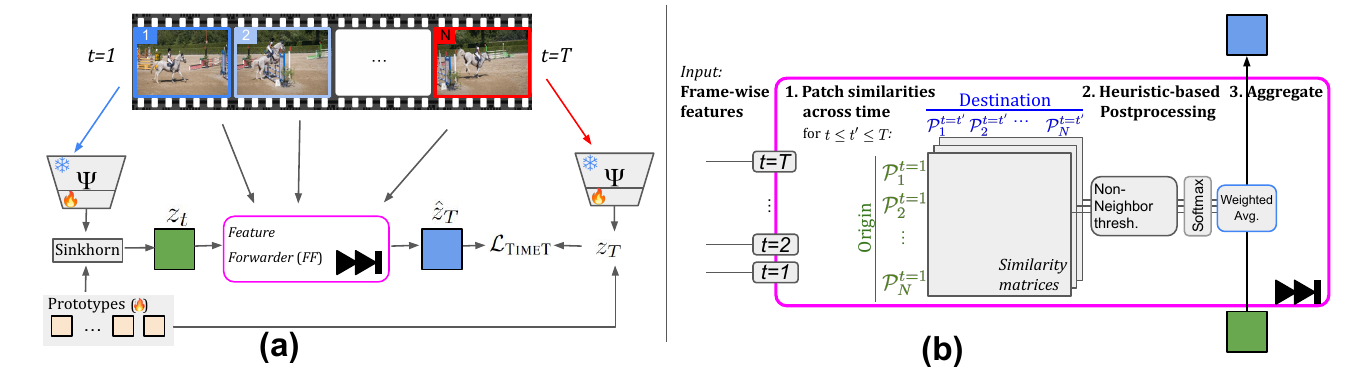}
    \caption{\textbf{\methodname~overview.} We conduct self-supervised video semantic segmentation by tuning image-pretrained models with temporal information from videos. \textbf{(a)}: Our general pipeline of adapting an image-pretrained ViT transformer on video data using our dense clustering loss. The encoder is specified by $\Psi$ and kept frozen except for the last two layers. Also, $z_t$ shows the output of Sinkhorn-Knopp algorithm, which is forwarded to time-step $T$ to be compared with the features that are obtained using the last time-step. \textbf{(b)}: Detailed view into our Feature-Forwarder module that is used for aligning cluster-maps from past frames to logit-features of future ones.}
    \label{fig:schematic}
\end{figure*}

%% file: 02_related_work.tex
\section{Related Works}
\paragraph{Dense self-supervised learning.}

These methods build upon image-level self-supervised representation learning by incorporating existing losses to enhance the spatial features, demonstrating a commendable advancement. 
DenseCL~\cite{wang2021dense} works on spatial features by constructing dense correspondences across views using the contrastive objective given in MoCo~\cite{he2020momentum}, while PixPro~\cite{xie2021propagate} utilizes the augmentation wrapper to get the spatial
correspondence of the pixel intersection between two views. Similarly, MaskContrast~\cite{van2021unsupervised}, Leopart~\cite{ziegler2022self}, DetCon~\cite{henaff2021efficient} and Odin~\cite{henaff2022object} also ensure spatial feature similarities via contrastive learning and spatial correspondences. SelfPatch~\cite{yun2022patch} treats the spatial neighbors of the patch as positive examples for learning more semantically meaningful relations among patches. Inspired by SelfPatch, ADCLR~\cite{zhang2023patchlevel} proposes patch-level contrasting via query crop and cross-attention mechanism.
Pursuing the same objective through none end-to-end approaches. Both~\cite{zadaianchuk2022unsupervised} and \cite{wang2023cut} propose an unsupervised salient object segmentation pipeline that extracts noisy object masks from the inputs and fine-tunes a specifically designed object segmentation head with several self-training steps to make an unsupervised semantic segmentation model. 
Unlike the existing image-based approaches, we propose a method that improves the dense prediction performance of a pretrained encoder by explicitly modeling the temporal dimension and learning from diverse natural dynamics and variations found in \textit{videos}.

\paragraph{Video to image knowledge transfer.} Learning from videos instead of images has recently received attention since they contain far more information than still images and hold the potential for learning rich representations of the visual world. To this end, several \textit{non-dense} works have been released~\cite{parthasarathy2022self, aubret2023time, hernandez2023visual}. VITO~\cite{parthasarathy2022self} shows that naively applying image domain self-supervised learning methods on videos can lead to a performance drop coming from the distribution shift, which by applying data processing techniques is relieved but not fully solved. Following the same way, \cite{hernandez2023visual} trains masked autoencoders~\cite{he2022masked} with contrastive learning~\cite{chen2020simple} on video datasets and shows a decent performance on both image and video tasks. While video to image knowledge transfer is the end goal of such methods and our paper; our dense task is different and is as of yet unaddressed.

\paragraph{Unsupervised video object segmentation.} Unsupervised video object segmentation methods do not require any manual annotations. However, they are only designed to tackle a \textit{foreground/background} segmentation task, which refers to the segmentation of the most prominent, general objects in video sequences~\cite{yang2021self, lu2020learning, ren2021reciprocal, araslanov2021dense, liu2021emergence}. For instance, \cite{yang2021self} uses an AutoEncoder-based architecture based on slot attention so that pixels that show the same motion flows based on optical-flow are grouped together. In this way, dominant moving objects can be detected and separated from the background. 
In our experiments, we benchmark two representative instances of this line of work, \ie,~\cite{yang2021self, araslanov2021dense}, to provide better insights into the differences of unsupervised object detection methods and our proposed benchmark for video semantic segmentation.

To conclude, we propose the first self-supervised video semantic segmentation method, which tunes image-trained models such as ~\cite{caron2021emerging, zhou2021ibot} based on the temporal information that exists in unlabeled videos to enhance their effectiveness in dense prediction tasks.

%% file: 03_method.tex
\section{Method}
\label{Sec. Method}

Our method works by densely clustering features in a manner that is consistent with time. At a high level, it works by transforming  features from a past time $t'{=}t$ to the current frame $t'{=}T$ and forcing these to be consistent with the currently observed ones at $t'{=}T$. 
The overall learning signal comes from the dense temporal clustering of forwarded features and allows ``timetuning'' image-pretrained models to learn from videos. Figure~\ref{fig:schematic} shows the proposed method's architecture. In the following, we describe each component in detail.

\subsection{Feature Forwarding}

The goal of this component, the \textit{feature-forwarder} (\textsc{FF}), is to propagate the dense features in time as accurately as possible, given some RGB input frames. 
More formally, let $f$ be the feature forwarding function, $I_t$ be a frame at time $t$ for a given video, and $z_t$ the features corresponding to $I_t$, then the task is to propagate $z_t$ to a future time $T$,
\begin{equation}
    f([I_t,\dots,I_T], z_t) \rightarrow \hat{z}_T.
    \label{eq:ff}
\end{equation}
Note that $\hat{z}_T$ and $z_T$ are not necessarily the same for $T > t$; since $\hat{z}_T$ represents an approximation of $z_T$ using the previous features. Obviously, the further $T$ is from $t$, the more variance is seen and therefore a better learning signal might be provided; however, for such larger intervals, the idea of feature forwarding encounters two main challenges:  error accumulation and object non-permanence.

\paragraph{Challenge 1: Error accumulation.}
While computing how features have changed simply based on the start and end frames is the most straight-forward (\ie, $I_{t}$ and $I_T$), it does not fully leverage the knowledge contained in the intermediate frames. 
Instead, it is common~\cite{jabri2020space, wang2019learning, li2019joint} to forward features for every time-step $\delta t$, \ie,
\begin{align}
    f([I_t,..,I_T], z_t) = \bigcirc_{t'=t+\delta t}^T f([I_{t'-\delta t},I_{t'}], z_{t}]),
    \label{eq:composing}
\end{align}
\noindent where $\bigcirc$ is a composition operator such as element-wise multiplication. 
However, this results in the accumulation of errors over time.

\paragraph{Challenge 2: Object non-permanence.}
One difficulty that arises when using videos as training data is the fact that objects sometimes simply disappear from the screen.
Compared to images, which carry a heavy photographer's bias, videos are more prone to natural variation due to camera and object motion resulting in temporary object \textit{non}-permanence {in a particular video clip.
A method that simply assumes an object (or a feature) has to be present in any given frame simply because it was present in the previous one might therefore easily fail when going towards videos that exist in-the-wild.
A case in point are simple occlusions that arise from an object being fully- or partially-covered by another one further in the background. Taking longer training clips has been attempted to resolve the issue. However, this increases the training complexity, especially for dense-tasks that necessitate a higher number of predictions for each input.

\vspace{1em}
\noindent \textit{We address these challenges using a novel, yet simple component, which we call the} \textbf{Stabilizing Feature-Forwarder.}
The first choice in designing this module is the function and data used for composing information across time. 
For instance, a simple approach would be to use pixel-wise optical flow and aggregate the flow at the feature level to predict how the features change over time.  However, as we also show in our experiments, optical flow is prone to accumulating error with time and also cannot easily recover from occlusions~\cite{Brox2011}. 
Instead, we utilize the fact that we wish to pretrain the visual encoder and \textit{recycle} part of it for the purpose of feature forwarding.
This not only speeds up the process, as no additional encoder/modality is required, but also produces a positive feedback loop of better forwarding and better feature learning.

Concretely, we first compute the L2-normalised dense features of a pretrained visual encoder $\Phi$ to compute spatial similarities across time (the $'$ in $t'$ left out for clarity): 
\begin{equation}
    F^t_{ij} = \langle \Phi(I_t)_i, \Phi(I_{T})_j \rangle/\tau, %
\end{equation}
which yields matrices $F^{\{t,\dots T-\delta t\}} \in R^{N \times N}$ with values between [0,1] indicating semantic similarities between the $N$ spatial features that are sharpened by a temperature $\tau$. 
Note that because we compare each frame with the final frame, the effect of object non-permanence can be minimized: even if the object is only present in every other frame, there is enough signal to forward its features.

Next to propagate from time $t$ to $T$, these similarities are stacked and normalised along time, as follows: 

\begin{align}
\tilde{F}^t_{ij} = \exp(\mathcal{N}(F^t_{ij}))/  \sum_{i, t'} \exp(\mathcal{N}(F^{t'}_{ij})),
\end{align}

\noindent where $ t \leq t'\leq T $ and $\mathcal{N}$ is a neighborhood thresholding method which forces short-term spatial-smoothness, \ie,
$\mathcal{N}(\tilde{F}^t_{ij}) {=} 0$ if $i$ is not within a local window around $j$ with the size $k$.
Finally, the propagated feature are computed $\hat{z}_{T}$ as:

\begin{align}
    \label{eq:final}
    \hat{z}_{T}(j) = \sum_{t', i} \tilde{F}^{t'}_{ij} \hat{z}_{t'}(i), \quad j \in \{1,..,N\}.
\end{align}
This means that for arriving at the target feature $\hat{z}_T$, we not only use the previous frame as the source, but instead use the past $(T-t)/\delta t$ frames and aggregate these.

\paragraph{Relation to mask-propagation methods.}
While previous methods such as DINO~\cite{caron2021emerging} and STC~\cite{jabri2020space} have utilized a similar technique for propagating ground-truth masks for evaluating, for example on the DAVIS dataset, there are three key differences to our forwarding method.
First, instead of propagating binary maps of foreground-vs-background, our feature-forwarder has as inputs soft, noisy and multi-label self-supervised segmentation maps, which require the forwarder to tolerate overlapping of different object probabilities throughout training.
Second, previous methods have used this approach mainly for inference; we, however, use this module as a trainable component and show how our loss improves this forwarder with training time.
Finally, we do not follow a typical re-normalizing step across feature dimensions (typically done after Eq. \eqref{eq:final}) as this harms the scale of logits that are being propagated and leads to heavily diluted target distributions.

\subsection{Self-supervised dense clustering}
While the Feature-Forwarder produces target features that include information about the dynamics with time, its computation utilizes $\Phi(I_T)$ and could lead to trivial solutions, \ie, $f([I_t,..,I_T], z_t)=\Phi(I_T)$. 
To counteract this, we propose a self-supervised clustering task across views in time.
For this, we utilize the basic online clustering algorithm based optimal-transport~\cite{cuturi2013lightspeed}, utilized in works such as SeLa~\cite{asano2020self}, SwAV~\cite{caron2020unsupervised} and DINOv2~\cite{oquab2023dinov2}.
In particular, let $\Psi$ be a visual encoder with a clustering head $g$ that yields a $K$ dimensional output,
then the clustering loss is given by:
\begin{equation}
    \mathcal{L}(x_i) = -\tilde{y}_i \log (g(\Psi(x)_i)),   
\end{equation}
which is a standard cross-entropy loss with regards to self-supervised pseudo-label $\tilde{y}_i$.
These labels, in turn, are generated by solving an entropy-regularised optimal transport problem on the batch $\mathcal{B}$~\cite{asano2020self}:
\begin{align}
      \min_{\tilde{y}} \langle \tilde{y}, - \log g(\Psi(x)) \rangle
  +  \frac{1}{\lambda} \operatorname{KL}(\tilde{y} \|rc^\top), \\
  \quad \text{with}\,\, r = \frac{1}{K}\cdot\mathbbm{1},\quad c = \frac{1}{|\mathcal{B}|}\cdot \mathbbm{1}.
\end{align}
Here $\lambda$ controls the entropy regularisation and $r,c$ are the marginals for the prototypes and the batch, respectively.
Note that solving this problem can be done extremely quickly on the GPU and yields soft pseudo-labels $\tilde{Y}$, such that $\text{argmax}(\tilde{Y}) = \tilde{y}$.

n, these might yield sufficiently similar semantics~\cite{feichtenhofer2021large} this is not true for dense tasks, as static features can only be assumed for videos where nothing is moving, which is rare.

\subsection{Overall~\methodabbrev~loss}
We combine the previous two modules to arrive at our full method (as shown in~\cref{fig:schematic}):
First, self-supervised Sinkhorn-Knopp clustering is conducted on early features $g(\Psi(I_t))$, yielding soft pseudo-labels $\text{SK}(g(\Psi(I_t))) {=} \tilde{Y}_t$. 
These are then forwarded in time using our Feature-Forwarder to arrive at dense targets $\textsc{FF}(\tilde{Y}_t)$, which are used in the final loss.
Compactly:
\begin{equation}
    \mathcal{L}_\mathrm{\methodabbrev}(I_T) = - \sum_{i,j} \textsc{FF}(\tilde{Y}_t) \log (g(\Psi(I_T))).  
\end{equation}

%% file: 04_experiments.tex
\section{Experiments}

\subsection{Setup}

\paragraph{Datasets.} 

We train our method and baselines on \textbf{\textit{YTVOS}}~\cite{xu2018youtube}, one of the largest video segmentation datasets available, and evaluate on \textbf{\textit{DAVIS17}}~\cite{pont20172017} and  {YTVOS}. For YTVOS the ground truth masks are only available for the first frames of the test and validation sets, and therefore, a fixed random 20\% of the training set is used for testing.  
For transfer learning experiments, we use the validation set of  \textbf{\textit{Pascal VOC 2012}}~\cite{everingham2010pascal}, As the dataset has been commonly used as a main reference for recent works in dense self-supervised image segmentation~\cite{ziegler2022self, van2021unsupervised, wang2021dense}. For completeness, we also report the performance on egocentric datasets that have less object-centric bias and are prevalent in real-world scenarios.  For egocentric experiments, we train on  \textbf{\textit{EPIC-KITCHENS-100}}~\cite{Damen2022RESCALING} and evaluate on \textbf{\textit{VISOR}}~\cite{VISOR2022}. Further details are provided in Appendix~\ref{Apdx_impl_details}.

\paragraph{Models and baselines.}
Currently, there is a lack of available unsupervised semantic video semantic segmentation methods. Nevertheless, we have included a comprehensive comparison of our method with state-of-the-art techniques in both image and video domains for unsupervised image semantic segmentation and unsupervised video object segmentation. To evaluate the image-based models, we utilized either official reported numbers or provided pretrained models of STEGO~\cite{hamilton2022unsupervised} and Leopart~\cite{ziegler2022self}. We have taken every measure to ensure fairness in our comparison. To do so, all the used pretrained models have the same pretraining dataset (ImageNet-1k), and the number of their backbone parameters is roughly similar. For those models that use extra datasets, for instance, Leopart, we select the pretrained backbones that closely match the specification of YTVOS training dataset. Additionally, we trained image-based models on the same video datasets, where we converted video data into their corresponding image data and reported their performance wherever possible. To ensure comprehensive analysis, we have included 
the recent concurrent work Flowdino~\cite{zhang2023boosting}, which utilizes optical flow to refine DINO features on unlabelled videos, in our comparison benchmarks as well. This is done only in cases where there is a shared experimental setup.

\paragraph{Evaluation procedure.}

Despite unsupervised object segmentation being a well-established evaluation in the image domain \cite{van2021unsupervised, ziegler2022self}, evaluating unsupervised video multi-label object segmentation is challenging due to the absence of an established evaluation protocol for video object semantic segmentation \textit{without} supervision. In this regard, we propose a set of evaluation protocols for unsupervised video multi-label object segmentation, which exploits existing video object segmentation datasets for evaluation purposes (see details in Appendix~\ref{Apdx_prtcl_details}). In our experiments, we discard any projection head used during training and evaluate ViT's spatial tokens directly, similar to ~\cite{van2021unsupervised, ji2019invariant, wang2021dense}, using four methods: classification with a linear head, classification with an FCN head, overclustering, and clustering with as many clusters as the number of ground truth objects. To fine-tune different heads on top of the frozen spatial tokens, we follow~\cite{van2021unsupervised}. For unsupervised semantic segmentation, we apply $K$-Means on all spatial tokens, where $K$ is chosen to be higher or equal to the ground-truth number of classes, following the common protocol in image clustering~\cite{ji2019invariant,van2020scan}. For grouping clusters to ground-truth classes, we match them either by pixel-wise precision or Hungarian matching on merged cluster maps~\cite{kuhn1955hungarian}. For video datasets, we allow the matching to be per-frame, per-video, and per-dataset. The evaluation metric is permutation-invariant, following~\cite{ji2019invariant}, and the results are averaged over five different seeds. Clustering evaluations are preferred as they require less supervision and fewer hyperparameters than training a linear classifier and operate directly on the learned embedding spaces. 
We report our results in mean Intersection over Union (mIoU) unless otherwise specified.

\begin{table*}[!tb]
\setlength{\tabcolsep}{3pt}
    \centering
    \begin{subtable}{.3\textwidth}
        \input{Tables/abl_num_prototypes}

    \end{subtable}
    \hfill
    \begin{subtable}{0.3\textwidth}
        \input{Tables/abl_interval}

    \end{subtable}
    \hfill
        \begin{subtable}{0.3\textwidth}
        \centering
        \input{Tables/abl_num_frames}

    \end{subtable}
    \caption{\textbf{Ablations of the key parameters of our method.} The model is trained for 12 epochs on Pascal VOC,  and results for unsupervised segmentation with clustering ($K{=}21$), overclustering  ($K{=}500$), and linear pixel-wise classification (LC) are shown. The stability of our method over a range of prototypes (50-300), inter-frame time intervals ($\delta T \in$[0.5s-1.0s]), and the number of training frames (4-8) at a fixed clip duration ($T$) shows the robustness of the method.} 
\end{table*}

\paragraph{Model training.} 
We train a ViT-Small with patch size 16 and initialized from ImageNet-pretrained DINO weights~\cite{caron2021emerging} using the proposed self-supervised loss function.
For the ablations and the main experiments, our models are trained for 12 and 30 epochs respectively. Further training details are provided in Appendix~\ref{Apdx_impl_details}. Code will be made available.

\subsection{Ablations}
We first examine the essential parameters of our method by training~\methodabbrev~on YTVOS and assessing its ability to perform semantic segmentation on Pascal VOC. In addition to presenting the results through clustering and overclustering, we also demonstrate linear classification outcomes.

\paragraph{Number of prototypes.}
We first ablate the influence of the number of prototypes on downstream segmentation performance. Results in Table~\ref{table:num_prototypes} indicate a sharp increase in performance with a rise in the number of prototypes, but once a sufficiently large number is reached, performance stabilizes. We observe peak performance with a moderate number of 200 prototypes.   The stability of our method over a range of prototypes (50-300) suggests that our approach is robust to the change of this parameter.

\paragraph{Understanding \textsc{FF}.}
Next, we  ablate the temporal dimensions that influence the working of the Feature-Forwarder. 
In Table~\ref{table:time_interval_effectiveness}, we vary the time interval $\delta T$ between frames whilst keeping a fixed number of four frames. We generally find an increase in performance when increasing the time-interval, with performance peaking at 0.5s. 
One of the most critical observations is the clear difference in clustering performance from 26.2\% to 42.8\%  between training on single frames (first row) \textit{vs.} training with multiple frames. 
This clearly indicates the significant benefit of utilizing temporal information from the Feature-Forwarder during training.

Next, in Table~\ref{table:num_frame_effectiveness} we vary the number of frames given a fixed clip duration of 2s.
Again, we find that a moderate number of frames between the source and target time is most favorable.
While additional frames do not degrade the performance much, they do add computational complexity, so we prefer using 4 frames in our main method.

\paragraph{Choice of propagation feature.}Finally, in Table~\ref{table:propagator_effect}, we vary the \textit{type of feature} that we forward to a future time.
In the first row, we report the performance when not using any feature forwarding – thus solely training on single-frame inputs (\ie, the same as row 1 in Table~\ref{table:num_frame_effectiveness}).
In the second row, ``Identity'' shows the performance that is obtained when the feature map from the source frame is simply forwarded to the future without any changes, which shows an increase in the performance compared to the first row. 
This shows that often, training videos are not very dynamic, and a static assumption can already lead to some gains. 
However, compared to forwarding the Sinkhorn-Knopp (SK) clustered features, these gains are small (+4.6\% \textit{vs.} + 17\%). 
Importantly, our Feature-Forwarder relies on forwarding SK-sharpened feature maps and does not work when simply forwarding the network's output logits $\Phi(x)$, followed by a clustering step. The reason for obtaining considerably lower numbers here can be traced back to using an un-entropy-regularized clustering algorithm. In this case, the logits tend to cause a few prototypes to dominate the cluster centers, exacerbated after the propagation, resulting in highly noisy and uninformative propagated logits.
This shows that careful design of the Feature-Forwarder is indeed required.

\input{Tables/abl_prop_feats}

\paragraph{Applying~\methodabbrev~to different models.} As shown in Table~\ref{table:DeTeFF_backbones}, our method is generalisable to different backbones and self-supervised learning initializations, enabling it to be an effective method to transfer the knowledge of the videos to images in an unsupervised way, which reduces the cost of labeling for different downstream tasks.

\input{Tables/DeTeFF_Backbones}

\input{Figs/Qualitative_pascal}

\subsection{Large-scale experiments}

\input{Tables/merged_tables}

\paragraph{Unsupervised video semantic segmentation.}
In this section, we train our~\methodabbrev~method on the YTVOS dataset and evaluate it for unsupervised video object segmentation on both DAVIS and YTVOS.
The results are shown in Table \ref{table:clustering_overclustering}.  It should be noted that the number of prototypes we analyzed in Table~\ref{table:num_prototypes} is not the number of clusters used to report accuracies for Table~\ref{table:clustering_overclustering}, as the actual number is usually unknown in the case of unsupervised learning. For the clustering experiment, we set K to the number of ground truth objects, which makes evaluation metrics easier to interpret and is commonly done in image clustering and segmentation. This evaluation is repeated but for larger numbers of clusters, in the ``overclustering'' scenario.  Note that in dense self-supervised learning, overclustering can be particularly important because the learned representations are used as a feature extractor for downstream tasks such as semantic segmentation or object detection. By using more fine-grained representations, the network may be able to extract more discriminative features such as object parts~\cite{ziegler2022self}, which can lead to better performances.

From Table~\ref{table:clustering_overclustering}, we observe a clear trend: our method, trained on YTVOS not only achieves superior performances on YTVOS, but also beats existing image-trained models on DAVIS by a large margin.
In particular, the state-of-the-art self-supervised clustering method, Leopart,  has 4\% lower performance on per dataset DAVIS evaluation for K=GT and 1.6\% lower performance on YTVOS. The gap even becomes larger when Leopart is trained on the same video dataset with 8\% to 16\% lower numbers in per dataset and per clip numbers across different datasets.
This means that when grouping objects of the same class over time and the whole dataset, the image-based self-supervised methods have trouble generalising to videos, where objects are not centered in the frame, can appear in varying poses, and are more difficult in general. 
A strong contender is Leopart, which matches the performance of DINO for per frame and per clip clustering, however, improves considerably with the per dataset clustering.
We attribute this to the fact that their dense learning objective improves mainly the generalisation capabilities of the learned representation, however falling short of generalising to temporal variations.
We make similar observations for overclustering, improving the DINO baseline which we utilize as initialisation by 12\% to 14\% across different datasets in the per dataset metric.

We conclude that the proposed method outperforms all other methods in learning robust and discriminative representations for dense video understanding, and image-based self-supervised learning may not have sufficient generalisation capacity.
This shows that if used right, time is a particularly strong inductive bias for self-supervised learning. Figure~\ref{fig:qualitative_pascal} shows segmentations returned by the proposed model trained on YTVOS and tested on YTVOS, DAVIS, and VISOR respectively. The proposed method groups objects accurately, and importantly, the frame segmentations are considerably consistent over time (see more examples in Appendix~\ref{Apdx_visualisations}). This highlights the importance and relevance of temporal fine-tuning, not just for higher accuracy, but crucially for consistency and
robustness.

\paragraph{Transfer from video training to images.}

Despite the common belief that features learned from videos perform worse when transferred to images~\cite{kipf2021conditional, gordon2020watching},  the results in Table~\ref{table:vary_backbone_pascal} demonstrate our method achieves high performance that match state-of-the-art methods directly trained on images, specifically for the K=500, FCN, and LC metrics.
We also compare our performance gains in videos against models designed for unsupervised image segmentation. The results show that models highly biased towards image datasets cannot provide the same performance when trained on videos, lagging behind our approach by 7\% to 30\%. Our results demonstrate that achieving high transfer performance on challenging tasks through video-based self-supervised learning is not only feasible but can also maintain high performance across modalities. These findings suggest that our method can drive further advances in self-supervised learning and inspire new directions for research in this field. Figure~\ref{fig:qualitative_pascal} shows the qualitative results on Pascal VOC. For more visualisations we refer to Appendix~\ref{Apdx_visualisations}.

\input{Tables/sota_pvoc}

\paragraph{Salient object segmentation.} In Table~\ref{table:jaccard_fg}, we compare foreground masks obtained with various DINO ViT based methods. We use the cluster-based foreground extraction protocol from~\cite{ziegler2022self} (details provided in Appendix~\ref{Apdx_exprmnt}).
First, we find that our method outperforms the original DINO attention maps consistently by  10-11\%. 
We also surpass the results of Leopart~\cite{ziegler2022self} and STEGO~\cite{hamilton2022unsupervised},  works that rely on the same pretrained backbone. Even for the evaluation on Pascal VOC, our method achieves higher performances despite the domain shift of having trained on videos.

\input{Tables/Jaccard_fg_score_comparison}

\paragraph{Generalisation to egocentric datasets.} We train our method on  EPIC-Kitchens-100~\cite{Damen2022RESCALING} and evaluate on VISOR~\cite{VISOR2022}. We use the official code to convert VISOR to a DAVIS-like structure, in which we can report our numbers for per frame and per clip evaluation. Note that, after conversion, the object IDs do not maintain global consistency in the whole dataset; therefore, we cannot report the per dataset number. As Table~\ref{table:visor} shows, \methodabbrev~ outperforms image-based competitors on egocentric datasets as well.

\input{Tables/VISOR}

\paragraph{Visual In-Context Learning evaluation.}

Here, we contrast our approach with a recently introduced benchmark, which assesses the in-context reasoning capabilities of models within the vision domain. Unlike linear or FCN classification methods, visual in-context learning evaluation obviates the need for fine-tuning or end-to-end training. Instead, it constructs validation segmentation maps using patch-level, feature-wise nearest neighbor similarities between validation images (referred to as "queries") and training samples (termed "keys"). This approach mirrors strategies in the NLP domain, aiming to evaluate the proficiency of models in learning tasks from limited examples presented as demonstrations. The results are shown by Table~\ref{table:Hummingbird}. Given that the vast majority of extant models in the domain utilize ViT-S16 as their backbone, we conducted a re-evaluation of their checkpoints to furnish a consistent and directly comparable evaluation table. Subsequently, we re-trained our model employing ViT-B16 and benchmarked it against other baselines as presented by \cite{balavzevic2023towards}. The results, as depicted in the table, indicate that even though our model was exclusively trained on videos, it registers performance metrics in line with Leopart~\cite{ziegler2022self}. Furthermore, it surpasses the results of the leading method, CrOC~\cite{stegmuller2023croc}, which was trained on images. Contrary to Hummingbird~\cite{balavzevic2023towards}, \methodabbrev~ is not custom-fitted to the specific evaluation setup and boasts superior computational efficiency. Notably, it requires only a single GPU for training, in contrast to the 16 TPUs demanded by Hummingbird. 

\input{Tables/Hummingbird}

%% file: Tables/abl_num_prototypes.tex
\centering
\caption{Ablating \# prototypes $K$.} %
\label{table:num_prototypes}
{\begin{tabular}{rrrr}
\toprule
\multirow{1}{*}{$P$}  & \multicolumn{1}{c}{LC} & \multicolumn{1}{c}{K=21} & K=500\\
\midrule

\multirow{1}{*}{10} & 37.5 & 6.1 & 24.1\\

\multirow{1}{*}{50} & 58.2 & 8.2 & 40.5\\

\multirow{1}{*}{200} & \textbf{59.7} & \textbf{9.2} & \textbf{42.8}\\
\multirow{1}{*}{300} & 59.4 & 9.0 & 42.3\\
\bottomrule
\end{tabular}}

%% file: Tables/abl_interval.tex
\centering
\caption{Ablating time interval $\delta T$.} %
\label{table:time_interval_effectiveness}
\setlength{\tabcolsep}{4pt}
    {\begin{tabular}{cr rrr}
    \toprule
    \# frames  & $\delta T$ & LC & K=21 & K=500\\
    \toprule
        \multirow{1}{*}{$1$ } & 0s   & 50.7         & 5.6 & 26.2\\
    \multirow{1}{*}{$4$ }     & 0.2s &   56.5       &  7.4   & 36.5\\
    \multirow{1}{*}{$4$ }     & 0.5s & \textbf{59.7} & \textbf{9.2} & \textbf{42.8} \\
    \multirow{1}{*}{$4$ }     & 1.0s &       57.1         & 8.3 & 38.1\\
    \bottomrule
    \end{tabular}}

%% file: Tables/abl_num_frames.tex
\centering
\caption{Ablating number of frames used.}
\label{table:num_frame_effectiveness}
    {\begin{tabular}{crrrr}
    \toprule
    \# frames & $T$ & LC & K=21 & K=500 \\
    \toprule
    1  & 0s   & 50.7         & 5.6 & 26.2\\
    2  & 2.0s & 52.6 & 6.2 & 37.1  \\
    4  & 2.0s & \textbf{59.7} & \textbf{9.2} & \textbf{42.8}\\
    8  & 2.0s & 59.7 & 9.0 & 42.3\\
    \bottomrule
    \end{tabular}}

%% file: Tables/abl_prop_feats.tex
\begin{table}[t]   
\centering
{\begin{tabular}{lccc}
\toprule
\texttt{FF} type& LC& K=21 & K=500\\
\midrule
None & 50.7 & 5.6 & 26.2\\
Identity & 55.3 & 7.4 & 36.1\\
$\Psi(x)$  & 53.1 & 6.6 & 35.5\\
$\mathbf{SK}(\Psi(x))$ & \textbf{59.7} & \textbf{9.2} & \textbf{42.8}\\
\bottomrule
\end{tabular}}
\caption{\textbf{Propagating different features in \textsc{FF}} on Pascal VOC. 
We find that our Sinkhorn-Knopp (SK) based module outperforms static training (`None' or `Identity'). Moreover, propagating logits $\Psi(x)$ does not work as well as SK-regularised features.}
\label{table:propagator_effect}
\end{table}

%% file: Tables/DeTeFF_Backbones.tex
\begin{table}[t]   
\centering
\resizebox{\linewidth}{!}{
\setlength{\tabcolsep}{0.32em}
\begin{tabular}{llccccc}
\toprule
&& \multicolumn{2}{c}{At Init} & & \multicolumn{2}{c}{\textbf{+\methodabbrev}}\\ 
\cmidrule{3-4}
\cmidrule{6-7}
Pretrain & Backbone & K=500 &  LC & & K=500 &  LC\\
\midrule
MSN~\cite{assran2022masked}  & ViT-S/16 & 26.0 & 55.4 & & 48.3{\color{blue}$_{\uparrow22.3}$} & 67.2{\color{blue}$_{\uparrow11.8}$}\\
iBOT~\cite{zhou2021ibot}       & ViT-S/16 & 32.1 & 62.1 & & 47.1{\color{blue}$_{\uparrow 15.0}$} & 67.1{\color{blue}$_{\uparrow \phantom{0}5.0}$}\\
DINO~\cite{caron2021emerging}  & ViT-S/8 & 22.5 & 55.8 & & 53.9{\color{blue}$_{\uparrow 31.4}$} & 69.5{\color{blue}$_{\uparrow 13.7}$}\\
DINO~\cite{caron2021emerging}  & ViT-B/16 & 28.9 & 59.1 & & 52.7{\color{blue}$_{\uparrow 23.8}$} & 70.2{\color{blue}$_{\uparrow 11.1}$}\\
\bottomrule
\end{tabular}
}
\caption{\textbf{Applying~\methodabbrev~to different pretrainings} on Pascal VOC. \methodabbrev~can boost ({\color{blue}$\uparrow$}) the performance of different backbones with different initialization by a considerable margin, showing the generality of our approach.}
\label{table:DeTeFF_backbones}
\end{table}

%% file: Figs/Qualitative_pascal.tex
\begin{figure*}[!htb]
    \centering
    \begin{center}
        \includegraphics[width=\linewidth, trim={1cm 23.9cm 1cm 3.0cm},clip]{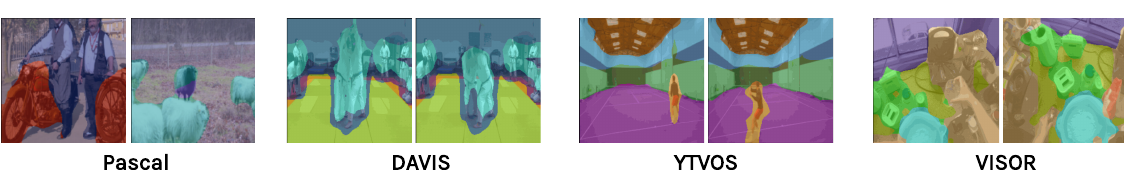}
    \end{center}
    \caption{\textbf{\methodabbrev~unsupervised segmentations.} 
    As we regularize DINO's backbone to be consistent across time on YTVOS, it obtains strong performance on both image and video segmentation datasets, yielding high class consistencies (indicated by the segmentation colors) and tight borders. We provide more 
    qualitative results in Appendix~\ref{Apdx_visualisations}.}
\label{fig:qualitative_pascal}
\end{figure*}

%% file: Tables/merged_tables.tex
\begin{table*}[h]   
\centering
{
\begin{tabular}{lrrrrrrrrrrrrrrrrr}
\toprule
& \multicolumn{7}{c}{\textbf{Clustering}} & & \multicolumn{7}{c}{\textbf{Overclustering}} \\
\cmidrule{2-8}
\cmidrule{10-16}
& \multicolumn{3}{c}{YTVOS} & & \multicolumn{3}{c}{DAVIS} & & \multicolumn{3}{c}{YTVOS} & & \multicolumn{3}{c}{DAVIS} \\
\cmidrule{2-4}
\cmidrule{6-8}
\cmidrule{10-12}
\cmidrule{14-16}
 & $\textit{F}$ & \textit{C} & \textit{D} & & $\textit{F}$ & \textit{C} & \textit{D} & & $\textit{F}$ & \textit{C} & \textit{D} & & $\textit{F}$ & \textit{C} & \textit{D} \\
\midrule
\textit{Trained on Images} & & & & & & & & & & & & \\
\quad {$\text{Resnet50}$} & 44.0 & 43.4 & 1.7 & & 39.3 & 37.4 & 4.2 &  & 55.6 & 52.8 & 3.1 & & 46.6 & 44.2 & 8.4 \\
\quad {SwAV~\cite{caron2020unsupervised}} & 39.5 & 38.2 & 3.2 & & 32.0 & 29.6 & 7.3 &  & 59.8 & 58.1 & 5.8 & & 50.5 & 50.1 & 25.7 \\
\quad {DINO~\cite{caron2021emerging}} & 39.1 & 37.9 & 1.9 & & 30.2 & 31.0 & 1.6 &  & 66.2 & 65.4 & 4.0 & & 56.9 & 54.9 & 17.9 \\
\quad {Leopart~\cite{ziegler2022self}} & 39.2 & 37.9 & 11.7 & & 30.3 & 30.2 & 16.5 &  & 64.5 & 62.8 & 15.5 & & 54.9 & 54.4 & 26.7 \\
\midrule
\textit{Trained on Videos} & & & & & & & & & & & & \\
\quad {STEGO*} & 41.5 & 40.3 & 2.0 & & 31.9 & 31.0 & 3.2 &  & 58.1 & 54.3 & 5.1 & & 47.6 & 46.3 & 10.4 \\
\quad {DINO*} & 37.2 & 36.1 & 1.2 & & 29.3 & 29.2 & 2.4 &  & 53.1 & 50.9 & 1.3 & & 45.4 & 44.0 & 8.6 \\
\quad {Leopart*} & 41.5 & 40.5 & 7.7 & & 37.5 & 36.5 & 12.6 &  & 60.8 & 59.8 & 6.8 & & 53.7 & 53.1 & 16.8 \\
\toprule
\quad {\methodabbrev (ours)} & \textbf{52.5} & \textbf{51.3} & \textbf{13.3} & & \textbf{53.7} & \textbf{53.0} & \textbf{20.5} & & \textbf{68.6} & \textbf{66.8} & \textbf{15.8} & & \textbf{59.8} & \textbf{61.5} & \textbf{31.8} \\
\bottomrule
\end{tabular}}
\caption{\textbf{Unsupervised video semantic segmentation results} in mIoU for clustering (K=GT) and overclustering. \methodabbrev~not only gets better numbers on YTVOS, but also achieves considerably better results on DAVIS. This shows the better quality of the learned features at transferability to other video datasets. The clip-length is set to 16 and 4 for DAVIS and YTVOS, respectively. For clustering, the Hungarian algorithm~\cite{kuhn1955hungarian} matches the unsupervised segmentation clusters (K) with the ground truth (GT) per frame (\textit{F}), clip (\textit{C}) or across the whole dataset (\textit{D}). For overclustering, we use K=10 for the frame-wise (\textit{F)} and clip-wise (\textit{C}) evaluations, and for dataset-wise (\textit{D}) evaluation, K=200 for DAVIS and K=500 for YTVOS. $^*$denotes finetuning pretrained models on the same video frames as input as our method. The matching protocol is greedy many-to-one, see Appendix \ref{Apdx_prtcl_details} for details. 
} 
\label{table:clustering_overclustering}
\end{table*}

%% file: Tables/sota_pvoc.tex
\setlength{\tabcolsep}{0.5em}
\begin{table}[t!]   
\centering
{\begin{tabular}{lrrrrrr}

\toprule & \multicolumn{4}{c}{Pascal VOC}\\
\cmidrule{2-5}
 & K=21 & K=500 & LC & FCN \\
\midrule
    \textit{Trained on Images}\\
    \quad ResNet-50                                           & 4.5               & 36.5& 53.8 & -              \\
    \quad DINO  \cite{caron2021emerging}                           & 5.5               & 17.4& 50.6 & 60.6                 \\
    \quad SwAV  \cite{caron2020unsupervised}                        & 11.6              & 35.7& 50.7  & -                \\
    \quad MaskContrast \cite{van2021unsupervised}                   & 35.0  &    45.4       &  49.2 & -         \\
    \quad DenseCL \cite{wang2021dense}                             & -  &    43.6       &  49.0 & 69.4         \\
    \quad STEGO~\cite{hamilton2022unsupervised} & 7.0 & 19.5 & 59.1 & 63.5\\
    \quad CrOC~\cite{stegmuller2023croc} & 20.6 & - & 61.6 & - \\
    \quad Leopart \cite{ziegler2022self}                           & \textbf{36.6}     & 50.5& \textbf{68.0} & 70.1          \\
    \midrule
    \textit{Trained on Videos}\\
    \quad STEGO* & 4.0 & 15.5 & 51.1 & 55.5\\
    \quad Leopart*                              & 14.9 & 21.2& 53.2 & 63.2          \\
    \quad Flowdino$^\dag$~\cite{zhang2023optical}      & - & - & 59.4 & -          \\
\midrule
    \quad \multirow{1}{*}{\methodabbrev~(ours)}  &   34.5  & \textbf{53.2} & \textbf{68.0} & \textbf{70.6}\\
\bottomrule
\end{tabular}}
\caption{\textbf{Transfer from video training to images.} 
Numbers taken from~\cite{ziegler2022self,stegmuller2023croc,zhang2023optical}. $^*$: finetuning pretrained models on the same video frames as input as our method. $\dag$: uses a 40\% larger superset of our train data. Details on the different datasets and methods are provided in the Appendix.
}
\label{table:vary_backbone_pascal}
\end{table}

%% file: Tables/Jaccard_fg_score_comparison.tex
\begin{table}[t!]   
\centering
{\begin{tabular}{lrrr}
\toprule 
  & Pascal VOC & DAVIS  & YTVOS\\
\midrule
    DINO  \cite{caron2021emerging}          & 52.1  & 34.5 & 32.1        \\
    Leopart \cite{ziegler2022self}          & 59.6  & 37.3 & 38.6         \\
    STEGO \cite{hamilton2022unsupervised}   & 49.1  & 30.4 & 32.1         \\
\midrule
\methodabbrev~(ours) &   \textbf{63.9}  & \textbf{44.5} & \textbf{43.5}\\
\bottomrule
\end{tabular}}
\caption{\textbf{Salient object segmentation.} We report performance using the Jaccard score~\cite{caron2021emerging} and use the official pretrained models for evaluation.}
\label{table:jaccard_fg}
\end{table}

%% file: Tables/VISOR.tex
\begin{table}[t]   
\centering
{\begin{tabular}{lcc}
\toprule
& \multicolumn{2}{c}{VISOR}\\
\cmidrule{2-3}
& $\textit{F}$ & \textit{C}\\
\midrule
DINO~\cite{caron2021emerging} & 24.8 & 18.7\\
Leopart~\cite{ziegler2022self} & 24.1 & 18.5\\
\toprule
\methodabbrev~(ours) & \textbf{26.5} & \textbf{21.5}\\
\bottomrule
\end{tabular}}
\caption{\textbf{Generalisation to egocentric datasets.} Unsupervised semantic segmentation results in mIoU for clustering (K=GT) on VISOR~\cite{VISOR2022} after training onn EPIC-Kitchens-100~\cite{Damen2022RESCALING}. The clip-length is set to 4. Our method gets better results on egocentric datasets as well.}
\label{table:visor}
\end{table}

%% file: Tables/Hummingbird.tex
\setlength{\tabcolsep}{0.5em}
\begin{table}[t!]   
\centering
{\begin{tabular}{lrrrr}

\toprule \\
 & Encoder & Params & mIoU\\
\midrule
    \textit{Trained on Images}\\
    \quad Supervised  & ViT-S16 & 21M & 35.1\\
    \quad MoCo-v3*~\cite{chen2021mocov3} & ViT-S16 & 21M & 19.5\\
    \quad DINO*  \cite{caron2021emerging} & ViT-S16 & 21M & 47.9\\
    \quad CrOC*~\cite{stegmuller2023croc} & ViT-S16 & 21M & 50.0\\
    \quad Leopart* \cite{ziegler2022self}  & ViT-S16 & 21M & 63.6\\
    \quad DINO  \cite{caron2021emerging} & ViT-B16 & 86M & 55.9\\
    \quad MoCo-v3~\cite{chen2021mocov3}  & ViT-B16 & 86M & 37.2\\
    \quad MAE~\cite{he2022masked} & ViT-B16 & 86M & 6.6\\
    \quad LOCA~\cite{caron2022location} & ViT-B16 & 86M & 57.5\\
    \quad Hummingbird~\cite{balavzevic2023towards} & ViT-B16 & 86M & \textbf{70.5}\\

    \midrule
    \textit{Trained on Videos}\\
    \quad \multirow{1}{*}{\methodabbrev~(ours)}  &   ViT-S16  & 21M & 61.6\\
    \quad \multirow{1}{*}{\methodabbrev~(ours)}  &   ViT-B16  & 86M & 65.5\\
\bottomrule
\end{tabular}}
\caption{\textbf{Visual In-Context Learning evaluation.} 
Numbers are taken from~\cite{balavzevic2023towards}. $^*$: numbers are produced by this paper. This appeoach is equivalent to a non-parametric nearest-neighbor based evaluation on Pascal VOC. The details of the evaluation benchmark can be found in the provided implementation codes.
}
\label{table:Hummingbird}
\end{table}

%% file: 05_conclusion.tex
\section{Conclusion}

This paper has aimed to learn dense representations that are learned from videos; yet can be generalised to the image domain as well. As video content is growing rapidly and contains more information compared to images, learning generalisable knowledge from them facilitates further scaling of self-supervised learning methods. To this effect, we have proposed a self-supervised clustering loss to encourage temporally consistent features between different frames of a video. To efficiently find corresponding views between clip frames, we have proposed to recycle pretrained transformer features to leverage their natural tracking ability at each clip. Our empirical results indicate that this method achieves significant gains on the challenging task of video object segmentation across three different evaluation protocols and three datasets. Moreover, by transferring the learned model to the image domain, we have demonstrated the generalisability of the learned features by surpassing or matching the state-of-the-art for unsupervised image segmentation. 

\section{Acknowledgement}
This work is financially supported by Qualcomm Technologies Inc., the University of Amsterdam and the allowance Top consortia for Knowledge and Innovation (TKIs) from the Netherlands Ministry of Economic Affairs and Climate Policy.

%% file: 99_appendix.tex
\appendix

{\Huge{Appendix}}

\section{Implementation Details}
\label{Apdx_impl_details}
Code is provided in the supplementary materials and will be open-sourced.

\paragraph{Training and evaluation datasets.} 

\paragraph{Video datasets.}
\textbf{\textit{DAVIS17}}~\cite{pont20172017} is designed for video object segmentation, comprising 150 videos, with 60 allocated for training, 30 for validation, and 60 for testing. Only the first frames of the test set have ground truth foreground masks, so the validation set is used for evaluation.
\textbf{\textit{YTVOS}}~\cite{xu2018youtube} is another dataset for video object segmentation and is significantly larger than DAVIS17. It consists of 4,453 videos that are annotated with 65 object categories. As with DAVIS17, ground truth masks are only available for the first frames of the test and validation sets, and therefore, a fixed 20\% of the training set is randomly sampled for the evaluation phase, details are provided in the supplementary material. Additionally, meta information is utilized to ensure objects in the same category have the same class id throughout the dataset for semantic, category-level assessments. Figure~ \ref{fig:appendix_class_distribution} shows the distribution of objects in YTVOS.

\paragraph{Image datasets.}

\textbf{\textit{Pascal VOC 2012}}~\cite{everingham2010pascal} is an object recognition dataset with 20 object categories and one background class. It includes pixel-level segmentation, bounding box, and object class annotations for each image, and has been extensively used as a benchmark for object detection, semantic segmentation, and classification tasks. The dataset is split into three subsets, with 1,464 images allocated for training, 1,449 for validation, and a private testing set. As the dataset has been commonly used as a main reference for recent works in dense self-supervised image segmentation~\cite{ziegler2022self, van2021unsupervised, wang2021dense}, we also use its validation set as one of the evaluation datasets.

\input{Figs/YTVOS_distribution}

\paragraph{Model training.}  We use batches of size 128 on 1 NVIDIA GeForce RTX 3090, and the optimizer is AdamW~\cite{loshchilov2017decoupled} with learning rate equal to 1e-4 for the projection head and the backbone’s learning rate is 1e-5. We freeze the backbone model except for the last two blocks for fine-tuning. Our model is implemented in torch~\cite{paszke2019pytorch}. We use Faiss~\cite{johnson2019billion} for K-Means clustering. We chose to train a ViT-Small~\cite{dosovitskiy2020image} image because it has roughly the same number of parameters as a ResNet-50 (21M vs. 23M). The projection head learning rate is 1e-4 and the backbone’s learning rate is 1e-5. The projection head consists of three linear layers with hidden dimensionality of 2048 and Gaussian error linear units as activation function~\cite{hendrycks2016gaussian}. We set the temperature to 0.1 and use Adam as an optimizer with a cosine weight decay schedule. The augmentations used are random color-jitter, Gaussian blur, grayscale, and random cropping.

\begin{lstlisting}[float, caption= \textbf{FF component}.
The pytorch implementation of \textbf{FF} is shown., label= FF_implementation, language=Python, backgroundcolor = \color{lightgray}]
# mm : torch.mm
# exp : torch.exp
# bmm : torch.bmm
# Reshape : In-place operation to change the input shape
# Normalize : torch.Normalize
# F[i] : Shows the ith feature map
# C-Map[i] : Shows the ith cluster map

prev_feat = []  # (nmb-context, dim, h*w)
prev_maps = []
For i in range(N-1):
    prev_feat.append(F[i])
    prev_maps.append(C-Map[i])
src_feat = Stack(prev_feat)
trgt_feat = F[N] # (1, dim, h*w)
trgt_feat= Normalize(trgt_feat, dim=1, p=2)
src_feat = Normalize(src_feat, dim=1, p=2)
aff = exp(bmm(trgt_feat, src_feat) / 0.1)
Reshape(aff, (nmb-context * h*w, h*w))
aff = aff / sum(aff, keepdim=True, axis=0)
aff = mask-neighborhood(aff) 
prev_maps = Stack(prev_maps)  # (nmb-context, C, h, w)
Reshape(prev_maps, (C, nmb-context*h*w))
trgt_cmap = mm(prev_maps, aff)

\end{lstlisting}

\paragraph{Evaluation details.} Since we evaluate the pre-GAP \textit{layer4} features or the spatial tokens, their output resolution does not match the mask resolution. To fix that, we bilinearly interpolate before applying the linear head; or directly interpolate the clustering results by nearest neighbor upsampling. For a fair comparison between ResNets and ViTs, we use dilated convolution in the last bottleneck layer of the ResNet such that the spatial resolution of both network architectures match (28x28 for 448x448 input images). All overclustering results were computed using downsampled 100x100 masks to speed up the Hungarian matching as we found that the results do not differ from using full-resolution masks.

\begin{algorithm}[t]
	\caption{Evaluation Pipeline Pseudocode. $M$ is the model, $C$ is the clustering algorithm, $\mathrm{MA}$ is the matching algorithm by which the clusters are scored, and $\mathrm{GT}$ is the given ground-truth.} 
	\label{algorithm:benchmark}
	\begin{algorithmic}[1]
	    \State input = input.reshape(bs * $n\textunderscore f$, c, h, w)
	    \State $\text{F}_{b}$ = M(input)
	    \State $\text{F}_{b}$ = F.reshape(bs, $n\textunderscore f$, num-patch, dim)
	    \State score\_list = []
	    \If{Per frame}
	        \For{$\text{F}_c$ In $\text{F}_{b}$}
	            \For{$\text{F}_f$ In $\text{F}_{c}$}
	                \State C\_Map = $C(\text{F}_f)$
	                \State score = $\mathrm{MA}(\text{C\_Map}, \mathrm{GT}_{f})$
	                \State score\_list.append(score)
	            \EndFor
	        \EndFor
  	    \ElsIf{Per clip}
        \For{$\text{F}_c$ In $\text{F}_{all}$}
            \State C\_Maps =  $C(\text{F}_c)$
            \State score = $\mathrm{MA}(\text{C\_Maps}, \mathrm{GT}_{c})$
            \State score\_list.append(score)
        \EndFor
  	    \ElsIf{Per dataset}
            \State C\_Maps =  $C(\text{F}_{b})$
            \State score = $\mathrm{MA}(\text{C\_Maps}, \mathrm{GT}_{b})$
            \State score\_list.append(score)
        \EndIf
        \State \textbf{return}(score\_list.mean())
	\end{algorithmic} 
\end{algorithm}

\section{Evaluation Protocol for Unsupervised Video Object Semantic Segmentation }

\label{Apdx_prtcl_details}

Here, we provide details for the evaluation protocols for unsupervised video multi-label object segmentation. To be consistent with the image domain~\cite{ziegler2022self}, a clustering algorithm is applied to the features extracted from frozen encoders to craft dense assignment maps of pseudo-labels. To produce scores, based on each evaluation protocol, the crafted maps are matched with the ground truth, and their MIOU is reported. Suppose the matching algorithm is specified by $\textit{M}(\text{labels}, \text{ground-truth})$, clustering algorithm by $K$, dataset features by $F \in R^{N \times n\textunderscore f \times d \times h \times w}$, where $N$, $n\textunderscore f$, $d$, $h$, and $w$  stand for dataset size, number of frames per clip, feature dimension, and feature spatial resolutions, respectively. 
 we introduce three evaluation protocols that are specific to the video domain.

\paragraph{Per frame evaluation (\textit{F}).}

\begin{align}
    & F_{\text{frame}}[i, j] = F[i, j] \\
    & \text{score} = \frac{1}{N \times n\textunderscore f}\sum_{i=1}^{N}\sum_{j=1}^{n\textunderscore f}\mathrm{MIOU}(M(\textit{K}(F_{\text{frame}}[i, j]), \mathrm{GT}[i,j]))
\end{align}

\noindent this measures a basic alignment of a given feature map with the ground-truth.

\paragraph{Per clip evaluation (\textit{C}).}

\begin{align}
    & F_{\text{clip}}[i] = (F[i, 1], \cdots, F[i, nf]) \\
    & \text{score} = \frac{1}{N}\sum_{i=1}^{N}\mathrm{MIOU}(M(\textit{K}(F_{\text{clip}}[i]), \mathrm{GT}[i]))
\end{align}

\noindent This evaluation tests whether the assigned pseudo-labels remain consistent over time for each clip.

\paragraph{Per dataset evaluation (\textit{D}).}

\begin{align}
    & F_{\text{dataset}} = (F[1, 1], \cdots, F[N, n\textunderscore f]) \\
    & \text{score} = \mathrm{MIOU}(M(\textit{K}(F_{\text{dataset}}), \mathrm{GT}))
\end{align}

\noindent This evaluation measures the most difficult ability of generating not only temporally stable features of objects across time but across videos.

\section{Additional Experiments}

\label{Apdx_exprmnt}

To provide a complete evaluation of our method compared to the baseline on Pascal VOC~\cite{everingham2010pascal}, we show the per-class performance in Figure~ \ref{fig:appendix_per_class}. As the figure shows, we improve the class "person" by more than 40\%, which could be beacuse of the high number of such objects in YTVOS as Figure \ref{fig:appendix_class_distribution} shows. The classes "cat" and "dog" also show a significant improvement since they are of the further dominant classes after "person".

\paragraph{Unsupervised video object segmentation and tracking.} 
Although such methods are designed for salient object detection or unsupervised mask propagation, and not specifically for unsupervised semantic segmentation, we evaluate their performance on our proposed evaluation protocols to highlight their strengths and limitations. The results are shown in Table \ref{table:trackingvsours}. As it is shown,~\methodabbrev~outperforms such methods on all the evaluation protocols by a margin between 12\% to 24\%. Not being specifically designed for semantic segmentation tasks may explain their inferior performance.

\paragraph{Comparing to unsupervised video object segmentation and tracking.} 
Although such methods are designed for salient object detection or unsupervised mask propagation, and not specifically for unsupervised semantic segmentation, we evaluate their performance on our proposed benchmark to highlight their strengths and limitations. We conducted a comprehensive comparison of our method with state-of-the-art techniques, such as Motion-Grp~\cite{yang2021self} and DUL~\cite{araslanov2021dense}, which aim to learn unsupervised features for propagating a given first frame's mask in test time or separating foreground from background using motion flows, as previously mentioned. To ensure consistency, we trained and tested all models on DAVIS, a widely used benchmark. Results in Table \ref{table:trackingvsours} demonstrate that our method outperforms these techniques across all reported metrics. It is worth noting that these methods were not specifically designed for semantic segmentation tasks, which may explain their inferior performance.

\input{Tables/CompareVSTracking_methods}

\paragraph{Analyzing per class results.} In Figure \ref{fig:appendix_per_class} and \ref{fig:k=gt_per_vlass} the per class improvements on Pascal VOC are reported. As the figures show, for the classes that make over 95\% of the number of objects existing in YTVOS, the performance almost always improves considerably. The reason that the class "bird" does not behave the same as the others might be due to the small size of this object  in YTVOS.

\input{Figs/DinoVsTimeT_k=gt}

\input{Figs/DINOvsDeTeFF_per_class}

\paragraph{Component Contributions.} In Table \ref{table:component_contribution}, we show the effect of using different components on the Pascal clustering results. As it is shown, \methodabbrev improves DINO~\cite{caron2021emerging} by 12\%. The results are improved by another 18\% after applying CBFE~\cite{ziegler2022self} to the features, showing high overclustering performance on this dataset. The number of clusters used for CBFE is 300 for this experiment.  

\input{Tables/Component_Contribution}

\paragraph{Comparing different propagators.}
Complementary to the previous ablations, in Table~\ref{table:propagator}, we conduct a detailed study on the performance of forwarding foreground masks using various temporal intervals and comparing the use of optical flow~\cite{sun2018pwc} with our Feature-Forwarder (\texttt{FF}).
We find that across all values of $\delta t$, \texttt{FF} outperforms flow by a large margin of +10\% or more.
This superiority has the added benefit of our \texttt{FF} module not having to expensively compute flow but instead reusing the activations that are processed for the clustering step.

\input{Tables/abl_prop_framerate}

\section{Additional Visualisations}

\label{Apdx_visualisations}

\input{Figs/qualitativ_davis}

\paragraph{Clustering with K=GT.} In Figure~ \ref{fig:appendix_k=21}, we show some qualitative results on Pascal VOC when K is set to the number of ground-truth objects, similar to Figure~\ref{fig:qualitative_pascal} in the paper. While the
method is trained on YTVOS with a temporal loss, it obtains strong performances on an image segmentation dataset yielding high class consistencies (indicated by the segmentation colors) and tight borders.

We also have provided further qualitative results with the same setting on DAVIS and YTVOS in Figure \ref{fig:qualitative_davis} and the attached HTML file. As the videos show, we get considerably more consistent and structured visualizations compared to DINO~\cite{caron2021emerging} and Leopart~\cite{ziegler2022self}. This further supports the effectiveness of temporal fine-tuning compared to the models solely trained on images.

\paragraph{Overclustering results by merging 500 clusters using ground-truth labels.} In the attached HTML file, we show the visualizations of our method on Pascal in the overslutering setting as well. As depicted, objects from different classes can be segmented precisely with different colors, showing that the learned patch features are semantic.

\input{Figs/Appendix_K=21_Visualizations}

\input{Figs/DINOvsOurs}

%% file: Figs/YTVOS_distribution.tex
\begin{figure}[!h]
    \centering
    \begin{center}
        \includegraphics[width=1\linewidth, trim={0.2cm 0.2cm 0.2cm 0.2cm},clip]{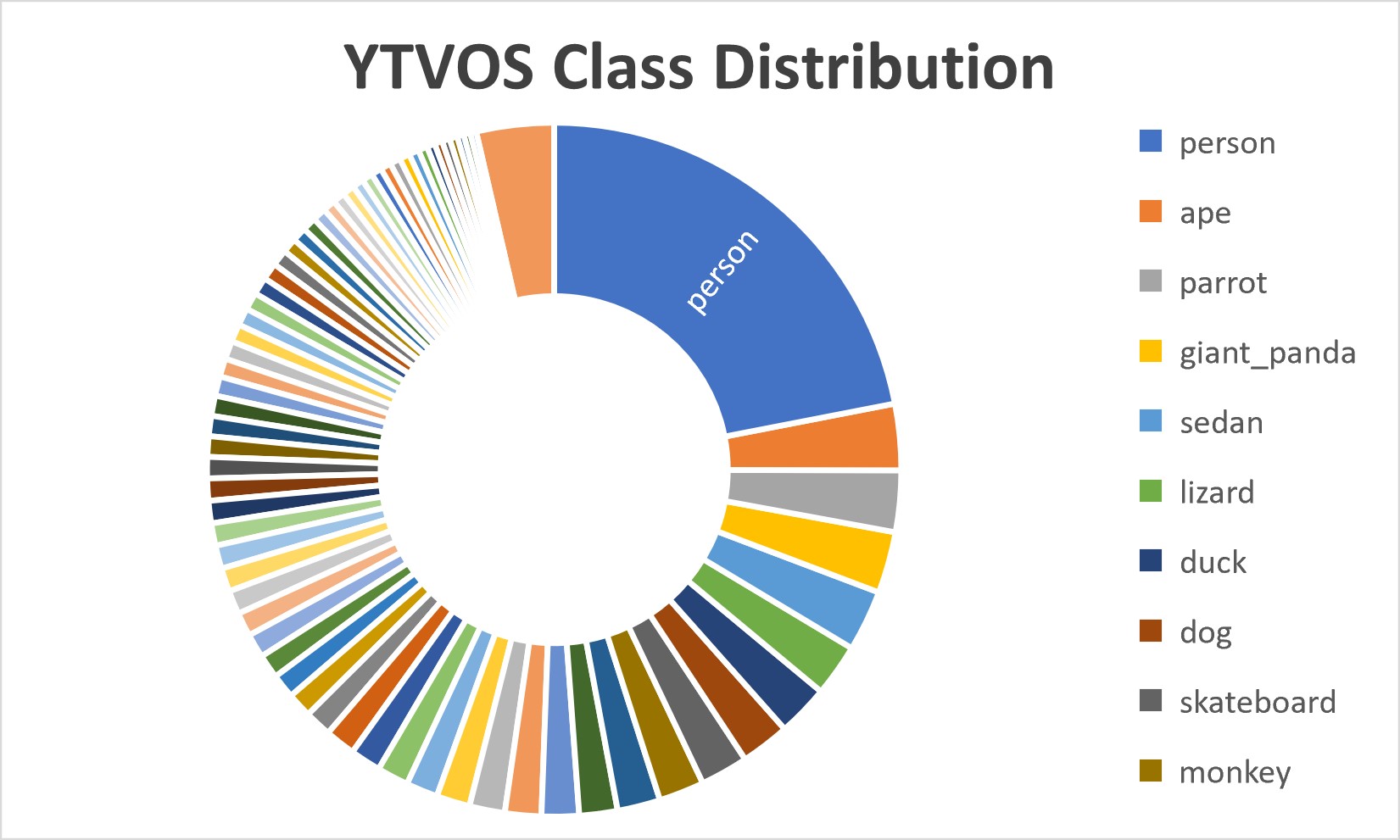}
    \end{center}
    \caption{The distribution of classes in YTVOS. Some of the more dominant classes are labeled.}
\label{fig:appendix_class_distribution}
\end{figure}

%% file: Tables/CompareVSTracking_methods.tex
\begin{table}[t]   
\setlength{\tabcolsep}{0.2em}
\aboverulesep = 0mm \belowrulesep = 0mm
\centering
{\begin{tabular}{lccc}
\toprule
Method & \textit{F} & \textit{C} & \textit{D}\\
\toprule

\multirow{1}{*}{DUL~\cite{araslanov2021dense}} & 28.2 & 27.4 & 2.4  \\
\multirow{1}{*}{Motion Grp~\cite{yang2021self}} & 32.0 & 30.7 & 1.5  \\

\toprule
\multirow{1}{*}{\methodabbrev~(ours)} & \textbf{56.5} & \textbf{55.5} & \textbf{14.1}\\

\bottomrule
\end{tabular}}
\caption{\textbf{Comparison to video unsupervised object segmentation methods.} 
Evaluation on DAVIS with K=GT.}
\label{table:trackingvsours}
\end{table}

%% file: Figs/DinoVsTimeT_k=gt.tex
\begin{figure*}[!t]
    \centering
    \begin{center}
        \includegraphics[width=0.9\linewidth, trim={0.2cm 0.2cm 0.2cm 0.2cm},clip]{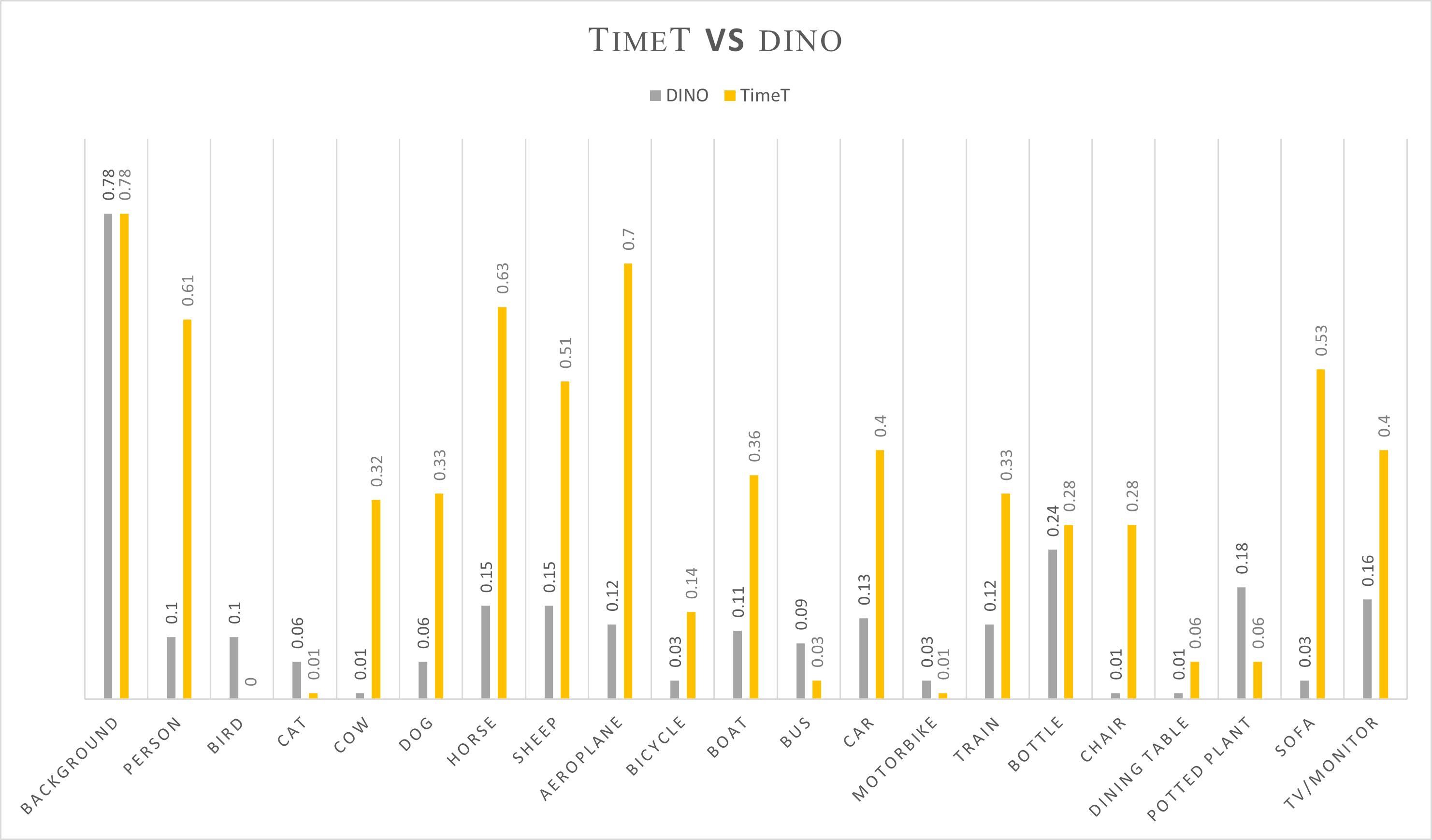}
    \end{center}
    \caption{The per class performance of DINO and \methodabbrev~ is shown for the clustering experiment with K=GT. Cluster-based foreground extraction~\cite{ziegler2022self} has been applied to both methods. As it is seen, this paper almost always improves the baseline performance for this evaluation as well. Pascal VOC is used in this experiment.}
\label{fig:k=gt_per_vlass}
\end{figure*}

%% file: Figs/DINOvsDeTeFF_per_class.tex
\begin{figure*}[!t]
    \centering
    \begin{center}
        \includegraphics[width=0.9\linewidth, trim={0.2cm 0.2cm 0.2cm 0.2cm},clip]{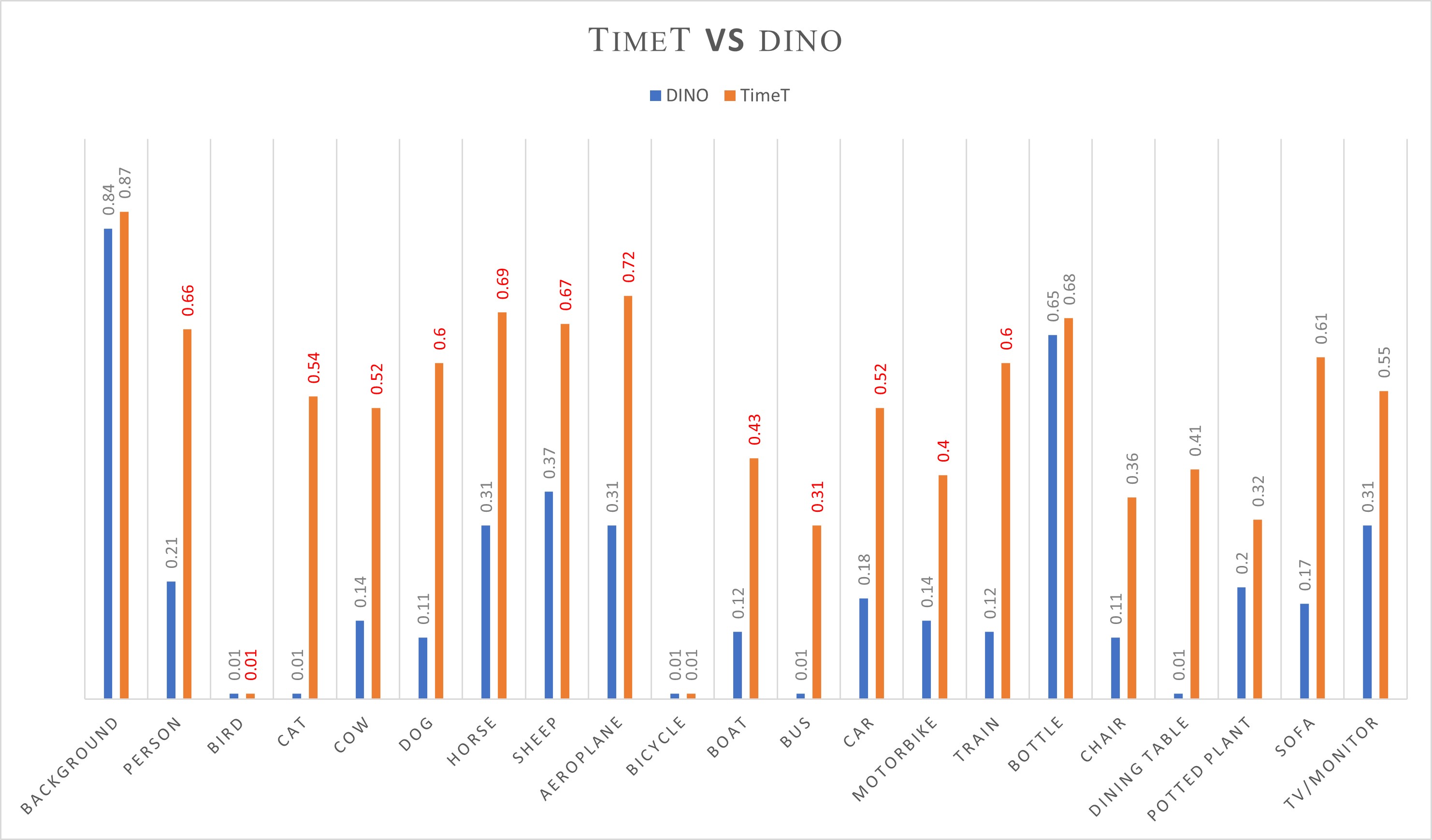}
    \end{center}
    \caption{The per class performance of DINO and \methodabbrev~ is shown for the overclustering experiment with K=500. As it is seen, this paper consistently improves the baseline performance. The numbers for the dominant shared classes between YTVOS and Pascal VOC are shown by the color red. }
\label{fig:appendix_per_class}
\end{figure*}

%% file: Tables/Component_Contribution.tex
\begin{table}[h]   
\centering
{\begin{tabular}{lc}
\toprule 
  & MIOU \\
\midrule
\toprule
    K=150 & 48.2 \\
\midrule
    DINO  & 4.6 \\
    $+\text{TimeT}$ & 16.5 \\
    $+\text{CBFE}$  & 34.5 \\
\midrule
\bottomrule
\end{tabular}}
\caption{\textbf{Component contributions.} We show
the gains that each individual component brings
for Pascal VOC segmentation and K=21.}
\label{table:component_contribution}
\end{table}

%% file: Tables/abl_prop_framerate.tex
\begin{table}[h]   
\centering
{\begin{tabular}{cccc}
\toprule
\multirow{1}{*}{$\delta t$}  & \multicolumn{1}{c}{$\text{GT}_0$} & \multicolumn{1}{c}{Flow} & \multicolumn{1}{c}{\texttt{FF}} \\
\midrule
\multirow{1}{*}{0.1s} & 26.4 & 29.8 & {39.8}\\
\multirow{1}{*}{0.2s} & 25.7 & 28.5 & \textbf{40.3}\\
\multirow{1}{*}{0.4s} & 24.5 & 26.1 & {40.1} \\
\multirow{1}{*}{0.8s} & 21.8 & 22.7 & {37.5} \\
\bottomrule

\end{tabular}}
\caption{Comparing \texttt{FF} with other forwarding methods. The numbers of reported on DAVIS validation set.} 
\label{table:propagator}
\end{table}

%% file: Figs/qualitativ_davis.tex
\begin{figure*}[!htb]
    \centering
    \begin{center}
        \includegraphics[width=0.95\linewidth, trim={0cm 4cm 1cm 1cm},clip]{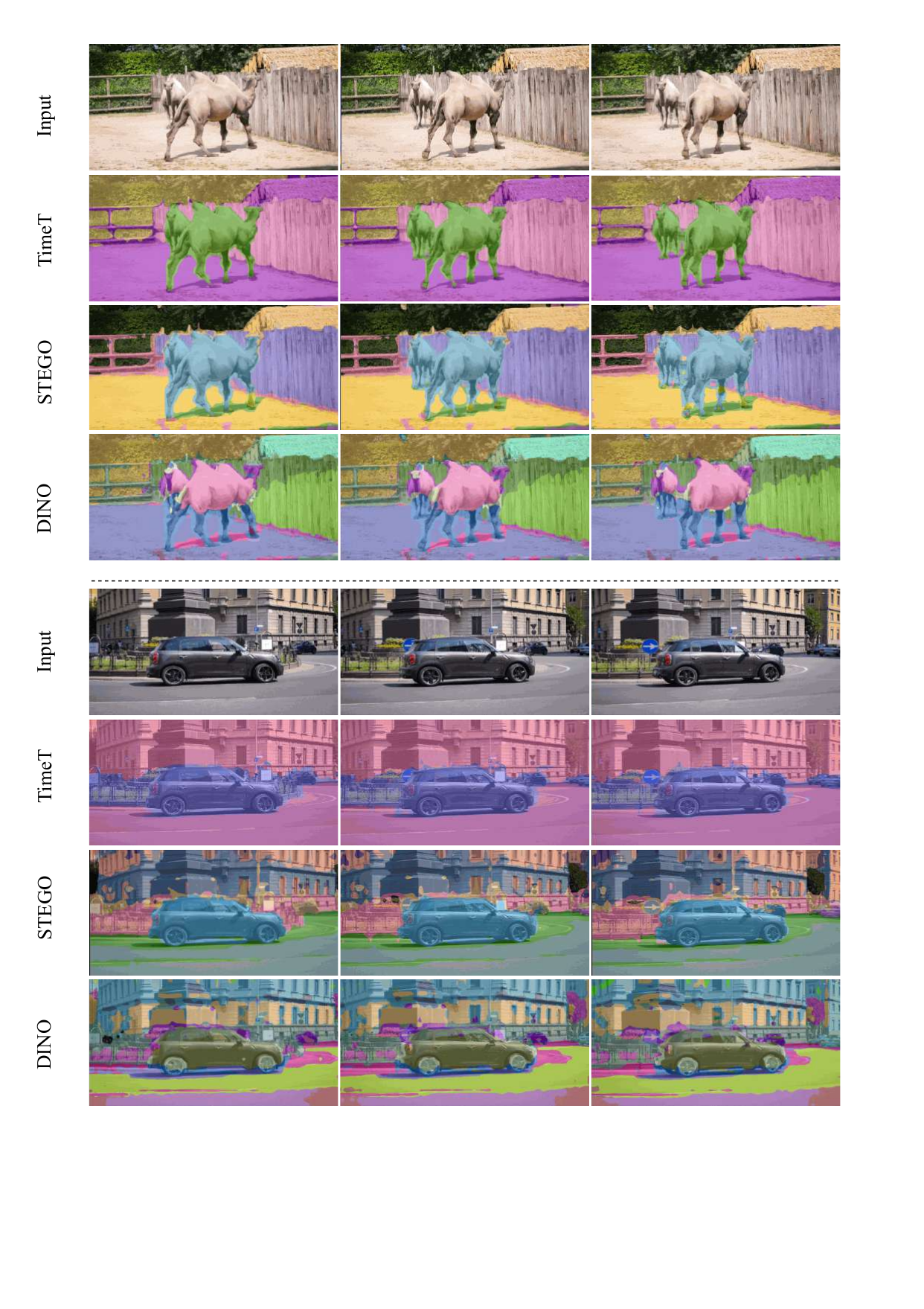}
    \end{center}
    \caption{\textbf{\textsc{\methodabbrev} segmentations on DAVIS with K=GT.} Here, we compare the performance of DINO, STEGO, and \methodabbrev~on the task of unsupervised video semantic segmentations. \methodabbrev ~ has a clear advantage over both DINO and STEGO in terms of providing tight segmentation boundaries and specifying different objects with different category IDs. Different colors in the figure specify different IDs.}
\label{fig:qualitative_davis}
\end{figure*}

%% file: Figs/Appendix_K=21_Visualizations.tex
\begin{figure*}[!t]
    \centering
    \begin{center}
        \includegraphics[width=0.9\linewidth, trim={2.5cm 5.5cm 2.5cm 1cm},clip]{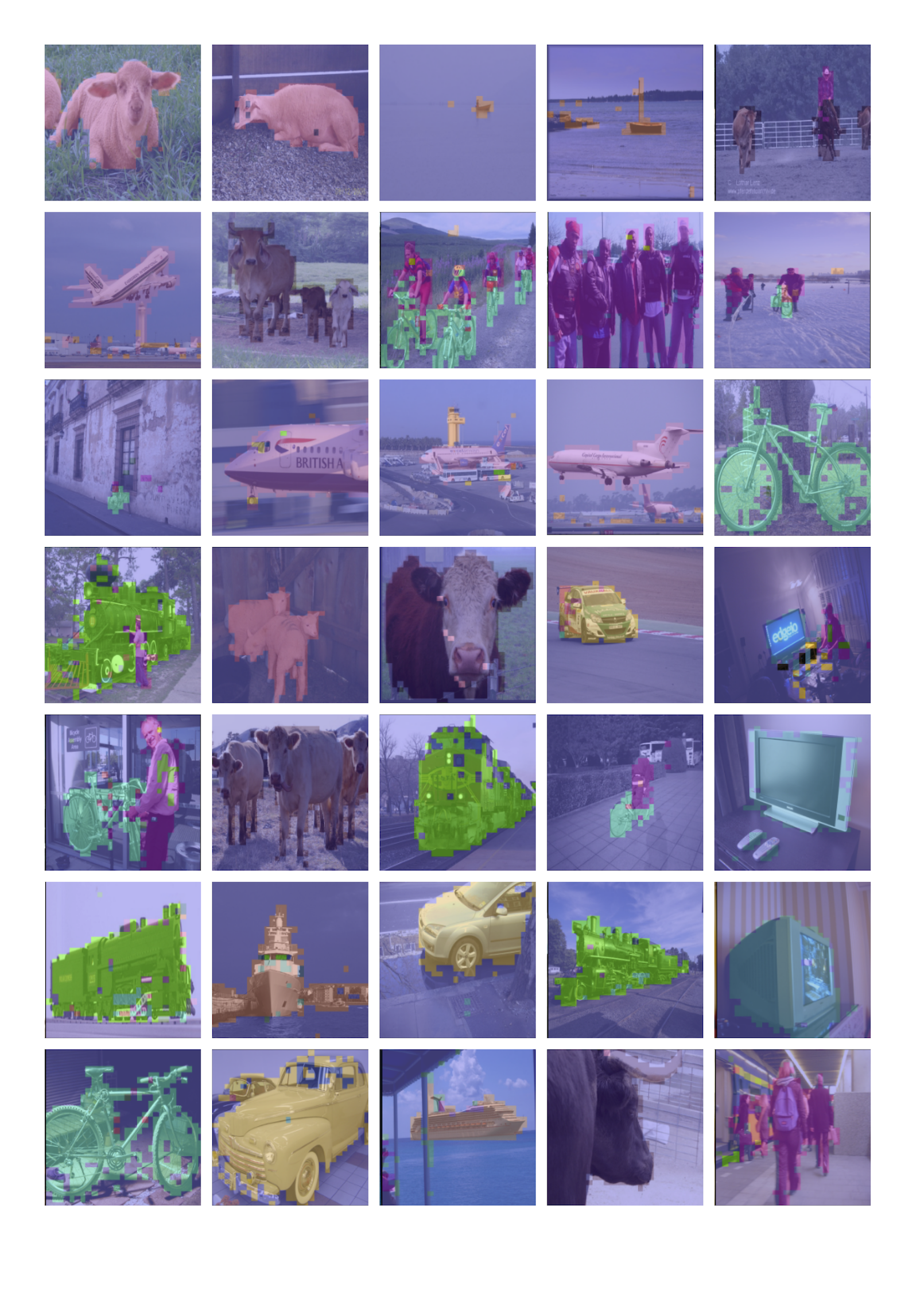}
    \end{center}
    \caption{\textbf{\textsc{\methodabbrev} segmentations on Pascal VOC with K=21.} 
    We use CBFE~\cite{ziegler2022self} to focus on the foreground objects. While the method is trained on YTVOS with a temporal loss, it obtains strong performances on an image segmentation dataset yielding high class consistencies (indicated by the segmentation colors) and tight borders.}
\label{fig:appendix_k=21}
\end{figure*}

%% file: Figs/DINOvsOurs.tex
\begin{figure}[!thb]
    \centering
    \includegraphics[width=\linewidth, trim={1cm 19.0cm 4cm 1cm},clip]{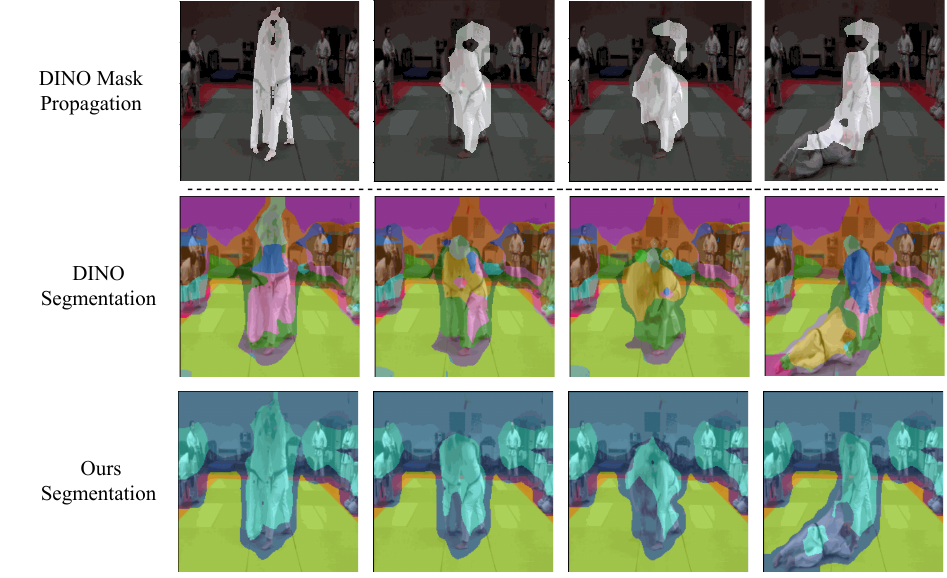}
    \caption{ DINO's features jump around across time, leading to inconsistent cluster maps. Our proposed \textsc{TimeT}-trained model observes more temporal consistency.}
\label{fig:TimeT-comparison}
\end{figure}